\documentclass[conference]{IEEEtran}
\IEEEoverridecommandlockouts
\usepackage{cite}
\usepackage{amsmath,amssymb,amsfonts,amsthm}
\usepackage{algorithmic}
\usepackage{graphicx}
\usepackage{textcomp}
\usepackage{comment}

\usepackage{balance} 
\usepackage{booktabs}
\usepackage{multicol}
\usepackage[ruled, vlined, linesnumbered]{algorithm2e}
\usepackage{float}
\usepackage{array}

\def\BibTeX{{\rm B\kern-.05em{\sc i\kern-.025em b}\kern-.08em
    T\kern-.1667em\lower.7ex\hbox{E}\kern-.125emX}}

\newtheorem{definition}{Definition}

\begin{document}

\title{
	Reward Shaping with Dynamic Trajectory Aggregation
}

\author{\IEEEauthorblockN{1\textsuperscript{st} Takato Okudo}
\IEEEauthorblockA{
	\textit{The Graduate University for Advanced Studies, SOKENDAI}\\
	Tokyo, Japna \\
	okudo@nii.ac.jp}
	\and
	\IEEEauthorblockN{2\textsuperscript{nd} Seiji Yamada}
	\IEEEauthorblockA{\textit{National Institute of Informatice, NII} \\
	Tokyo, Japan \\
	seiji@nii.ac.jp}
}

\maketitle

\begin{abstract}
Reinforcement learning, which acquires a policy maximizing long-term rewards, has been actively studied. Unfortunately, this learning type is too slow and difficult to use in practical situations because the state-action space becomes huge in real environments. The essential factor for learning efficiency is rewards. Potential-based reward shaping is a basic method for enriching rewards. This method is required to define a specific real-value function called a ``potential function" for every domain. It is often difficult to represent the potential function directly. SARSA-RS learns the potential function and acquires it. However, SARSA-RS can only be applied to the simple environment. The bottleneck of this method is the aggregation of states to make abstract states since it is almost impossible for designers to build an aggregation function for all states. We propose a trajectory aggregation that uses subgoal series. This method dynamically aggregates states in an episode during trial and error with only the subgoal series and subgoal identification function. It makes designer effort minimal and the application to environments with high-dimensional observations possible. We obtained subgoal series from participants for experiments. We conducted the experiments in three domains, four-rooms(discrete states and discrete actions), pinball(continuous and discrete), and picking(both continuous). We compared our method with a baseline reinforcement learning algorithm and other subgoal-based methods, including random subgoal and naive subgoal-based reward shaping. As a result, our reward shaping outperformed all other methods in learning efficiency.
\end{abstract}


\section{Introduction}
Reinforcement learning(RL) can acquire a policy maximizing long-term rewards in an environment. Designers do not need to specify how to achieve a goal; they only need to specify what a learning agent should achieve with a reward function. A reinforcement learning agent performs both exploration and exploitation to find how to achieve a goal by itself. It is common for the state-action space to be quite large in a real environment like robotics. As the state-action space becomes larger, the number of iterations exponentially increases to learn the optimal policies, and the learning becomes too slow to obtain optimal policies in a realistic amount of time. Since a human could have knowledge that would be helpful to such an agent in some cases, a promising approach is utilizing human knowledge~\cite{ijcai2018-817, DBLP:conf/atal/HarutyunyanBVN15a, 8708686}. \par 
The reward function is the most related to learning efficiency. Most difficult tasks in RL have a sparse reward function\cite{10.5555/3327144.3327216}. The agent is not able to evaluate its policy due to it and to learn the policy. In contrast, learning speeds up when the reward function is dense. Inverse reinforcement learning~(IRL)~\cite{Ng:2000:AIR:645529.657801, 10.1145/1015330.1015430} is the most popular method for enriching the reward function. IRL uses an optimal policy to generate a dense reward function. Recent studies have utilized optimal trajectories~\cite{10.5555/1620270.1620297, NIPS2016_6391}. There is the question of the cost of the teacher in providing trajectories or policies. Humans sometimes have difficulty providing these because of the skills they may or they may not have. In particular, in a robotics task, humans are required to have robot-handling skills and knowledge on the optimal trajectory. Another approach is reward shaping. This method expands the original environmental reward function. Potential-based reward shaping is able to add external rewards while keeping the optimal policy of the environment~\cite{Ng+HR:1999}. It is calculated as the difference between the real-number functions~(potential function) of the previous and current state. In \cite{Ng+HR:1999}, it was mentioned that learning sped up with a learned value function used as the potential function. Since a policy learned with a potential-based reward shaping is equivalent to that with $Q$-value initialization with the potential function~\cite{DBLP:journals/corr/abs-1106-5267}, using the learned value function for the potential function is equivalent to initializing the value function with the learned value function. Therefore, learning is jump-started. To use potential-based reward shaping, we need to define the potential function. It is often very difficult to represent the potential function directly. To solve this problem, SARSA-RS acquires it in learning\cite{Grzes2010}. A designer provides the aggregation function of states before learning, and SARSA-RS builds a value function over abstract state space as the potential function. The propagation of the reward over value function accelerates because the abstract state space is smaller than the original state space, and the agent learns the policy faster. Devlin and Kudenko proved that time-varied reward shaping keeps a policy invariant~\cite{Devlin:2012:DPR:2343576.2343638}. The result has made clear that a policy learned with SARSA-RS is equivalent to the original policy.
However, it is very difficult to define the aggregation function of states when the task has a high-dimensional state space. We propose a subgoal-based trajectory aggregation method. The designer defines only the subgoal identification function to apply SARSA-RS to a reinforcement learning algorithm. Since it is easier for the designer to make a similarity function than an aggregation function \cite{ijcai2017-534}, our method can keep designer effort minimal. Moreover, a non-expert may enhance the reinforcement learning algorithm if an identification function exists. This might be related to interactive machine learning\cite{Amershi_Cakmak_Knox_Kulesza_2014}. Providing subgoals is sometimes easier than trajectories because it does not require handling skills but only task decomposition skills. 

\section{Related Work}

The landmark-based reward shaping of Demir et al.~\cite{Demir2019} is the closest to our method. The method shapes only rewards on a landmark using a value function. Their study focused on a POMDP environment, and landmarks automatically become abstract states. We focus on an MDP environment, and we propose an aggregation function. We acquire subgoals from human participants, and we apply our method to a task with high-dimensional observations. Potential-based advice is reward shaping for states and actions\cite{Wiewiora2003}. The method shapes the q-value function directly for a state and an action, and it makes it easy for a human to advice to an agent regarding whether an action in an arbitrary state is better or not. Subgoals show what ought to be achieved on the trajectory to a goal. We adopted the shaping of a state value function.
Harutyunyan et al.~\cite{Harutyunyan2015} has shown that the q-values learned by arbitrary rewards can be used for the potential-based advice. The method mainly assumes that a teacher negates the agent's action selection. The method uses failures in the trial and errors. In contrast, our method uses successes. \par
In the field of interactive reinforcement learning, a learning agent interacts with a human trainer as well as the environment\cite{doi:10.1080/09540091.2018.1443318}.The TAMER framework is a typical interactive framework for reinforcement learning\cite{KCAP09-knox}. The human trainer observes the agent's actions and provides binary feedback during learning. Since humans often do not have programming skills and knowledge on algorithms, the method relaxes the requirements to be a trainer. We aim for fewer trainer requirements, and we use a GUI on a web system in experiments with navigation tasks. \par
Our method is similar to hierarchical reinforcement learning(HRL) in a hierarchy. The option framework is the major in the field of HRL. The framework of Sutton et al.~\cite{Sutton:1999:MSF:319103.319108} was able to transfer learned policies in an option. An option consists of an initiation set, an intra-option policy, and a termination function. An option expresses a combination of a subtask and a policy for it. The termination function takes on the role of subgoal because it terminates an option and triggers the switching to another option. Recent methods have found good subgoals for a learner simultaneously with policy learning~\cite{Bacon:2017:OA:3298483.3298491, 10.5555/3305890.3306047}. The differences with our method are whether the policy is over abstract states or not and whether rewards are generated. The framework intends to recycle a learned policy, but our method focuses on improving learning efficiency. \par
Reward shaping in HRL has been studied in~\cite{Gao2015, Li2019}. Gao et al.~\cite{Gao2015} has shown that potential-based reward shaping remains policy invariant to the MAX-Q algorithm. Designing potentials every level is laborious work. We use a single high-level value function as a potential, which reduces the design load. Li et al.~\cite{Li2019} incorporated an advantage function in high-level state-action space into reward shaping. Their approach is similar to ours in the utilization of a high-level value function, but it does not incorporate external knowledge into their algorithm. The reward shaping method in~\cite{Paul2019} utilized subgoals that are automatically discovered with expert trajectories. The potentials generated every subgoal are different. The value of a potential is fixed and not learned. Our method learns the value of a potential.

\section{Preliminaries and Notation}
A Markov Decision Process consists of a set of states $S$, a set of actions $A$, a transition function $T: S\times A \rightarrow (S' \rightarrow [0,1])$, and a reward function $R: S\times A \rightarrow \mathcal{R}$. A policy is a probability distribution over actions conditioned on states, $\pi:S\times A \rightarrow [0, 1]$. In a discounted fashion, the value function of a state~$s$ under a policy $\pi$, denoted $v_\pi(s)$, is $V_\pi = \mathbf{E}_\pi \left[ \sum_{i=0}^\infty \gamma^i r_{t+1} | s_0=s \right]$. Its action-value function is $Q(s,a) = \mathbf{E}_\pi \left[ \sum_{i=0}^\infty \gamma^i r_{t+1} | s_0=s, a_0=a \right]$, where $\gamma$ is a discount factor.
\subsection{Potential-Based Reward Shaping}
Potential-based reward shaping~\cite{Ng+HR:1999,DBLP:journals/corr/abs-1106-5267} is an effective method for keeping an original optimal policy $\pi$ in an environment with an additional reward function $F$. If the potential-based shaping function $F$ is formed as:
\begin{eqnarray}
	F(s_t,s_{t+1}) = \gamma \Phi(s_{t+1}) - \Phi(s_t)
\end{eqnarray}
, it is guaranteed that policies$\pi$ in MDP $M=(S,A,T, \gamma, R)$ are consistent with those $\pi'$ in MDP $M'=(S,A,T,\gamma, R+F)$. Note that $s_t \in S-\{s_0\}$ and $s_{t+1} \in S$. $s_0$ is an absorbing state, so the MDP ``stops" after a transition into $s_0$. $\Phi$ is known as the potential function. $\Phi$ should be a real-value function such as $\Phi: S \rightarrow \mathbf{R}$. For better understanding, we use the example of Q-learning with potential-based reward shaping. The learning rule is formally written as
\begin{gather}
	Q(s_t,a_t) \leftarrow Q(s_t, a_t) +\alpha\delta_{TD} \\
	\delta_{TD} = r_t + F(s_t, s_{t+1}) + \max_{a'} Q(s_{t+1}, a') - Q(s_t,a_t)
\end{gather}

, where $\alpha$ is a learning rate. We need to define an appropriate $\Phi$ for every domain. There is the problem of how to define $\Phi$ to accelerate learning. The study of~\cite{DBLP:journals/corr/abs-1106-5267} has shown that learning with potential-based reward shaping is equivalent to $Q$-value initialization with the potential function$\Phi$ before learning. The result has made clear that $\Phi(s) = V^*(s) = \max_{a'} Q^*(s, a')$ is the best way to accelerate learning. We cannot know $V^*(s)$ before learning since we acquire $V^*(s)$ after learning. This suggests that we can accelerate the learning if there is a value function learned faster than another with the same rewards.

\subsection{SARSA-RS}
Grzes et al.~\cite{Grzes2008, Grzes2010} proposed a method that learns a potential function $\Phi$ during the learning of a policy $\pi$, called  ``SARSA-RS". The method solved the problem of the design of an appropriate potential function for a domain being too difficult and time-consuming. We define $Z$ as a set of abstract states. The method builds a value function over $Z$ and uses it as $\Phi$
\begin{eqnarray}
	\Phi(s) = V(g(s)) = V(z)
\end{eqnarray}
, where $g$ is an aggregation function, $g: S \rightarrow Z$. The function $g$ is pre-defined. The potential-based shaping function over SARSA-RS is written as follows.
\begin{eqnarray}
	F(z_t, z_{t+1}) = \gamma V(z_{t+1}) - V(z_t)
\end{eqnarray}
The method learns the value function $V(z)$ during policy learning as:
\begin{eqnarray}
	V(z_t) \leftarrow V(z_t) + \alpha \left( r_h + \gamma^k V(z_{t+1}) - V(z_t) \right)
\end{eqnarray}
, where $r_h$ is the transformation function from MDP rewards into SMDP rewards, and $k$ is the duration between $z_{t}$ and $z_{t+1}$. The potential function changes dynamically during learning, and the equivalency of the potential-based reward shaping cannot is applied because it depends on the time in addition to the state.
Since Devlin and Kudenko have shown that a shaped policy is equivalent to a non-shaped one, when the potential function changes dynamically during learning\cite{Devlin:2012:DPR:2343576.2343638}, SARSA-RS keeps the learned policy original. We omit the time argument in the following section to simplify the expression. The size of $Z$ is smaller than $S$ thanks to the aggregation of states. Therefore, the propagation of environmental rewards is faster, and the policy learning with SARSA-RS is also faster. As mentioned above, the method requires the pre-defined aggregation function $g$. In an environment of high-dimensional observations, it is almost impossible to make an aggregation function.
\section{Reward Shaping with Subgoal-Based Aggregation}
\label{sec:reward-shaping}
We propose a method of aggregation from states into an abstract states. The method basically follows SARSA-RS. We use a pre-defined subgoal series and aggregate episodes dynamically into abstract states during learning with it. 

\subsection{Subgoal}
We define a subgoal as follows
\begin{definition}
	A state $s$ is a {\it subgoal} if $s$ is a goal in one of the sub-tasks decomposed from a task.
\end{definition}
In the option framework, the subgoal is the goal of a sub-task, and it is expressed as a termination function\cite{Sutton:1999:MSF:319103.319108}. Many studies on the option framework have developed automatic subgoal discovery~\cite{Bacon:2017:OA:3298483.3298491}. We aim to incorporate human subgoal knowledge into the reinforcement learning algorithm with less human effort required. The property of a subgoal might be a part of the optimal trajectories because a human should decompose a task to achieve the goal. We acquire a subgoal series and incorporate the subgoals into our method in experiments. The subgoal series is written formally as $(SG, \prec)$. $SG$ is a set of subgoals and a sub-set of $S$. There are two types of subgoal series, totally ordered and partially ordered. With totally ordered subgoals, a subgoal series is deterministically determined at any subgoal. In contrast, partially ordered subgoals have several transitions to the subgoal series from a subgoal. We used only the totally ordered subgoal series in this paper, but both types of ordered subgoals are available for our proposed reward shaping. Since an agent needs to achieve a subgoal only once, the transition between subgoals is unidirectional. 

\subsection{Subgoal-Based Dynamic Trajectory Aggregation}
We propose a method of aggregating trajectories dynamically into abstract states using subgoal series. The method makes the SARSA-RS method available for environments of high-dimensional observations thanks to less effort being required from designers. The method requires only a subgoal series consisting of several states instead of all states. In this section, we assume that the subgoal series $(SG, \prec)$ is pre-defined, and $(SG, \prec) = \left\{sg_0 \prec sg_1 \prec \cdots \prec sg_n \right\}$. The method basically follows SARSA-RS, and the difference is mainly the aggregation function $g$ and minorly the accumulated rewards.
\subsubsection{Dynamic Trajectory Aggregation}
\begin{figure}[tb]
	\centering
	\includegraphics[width=\linewidth]{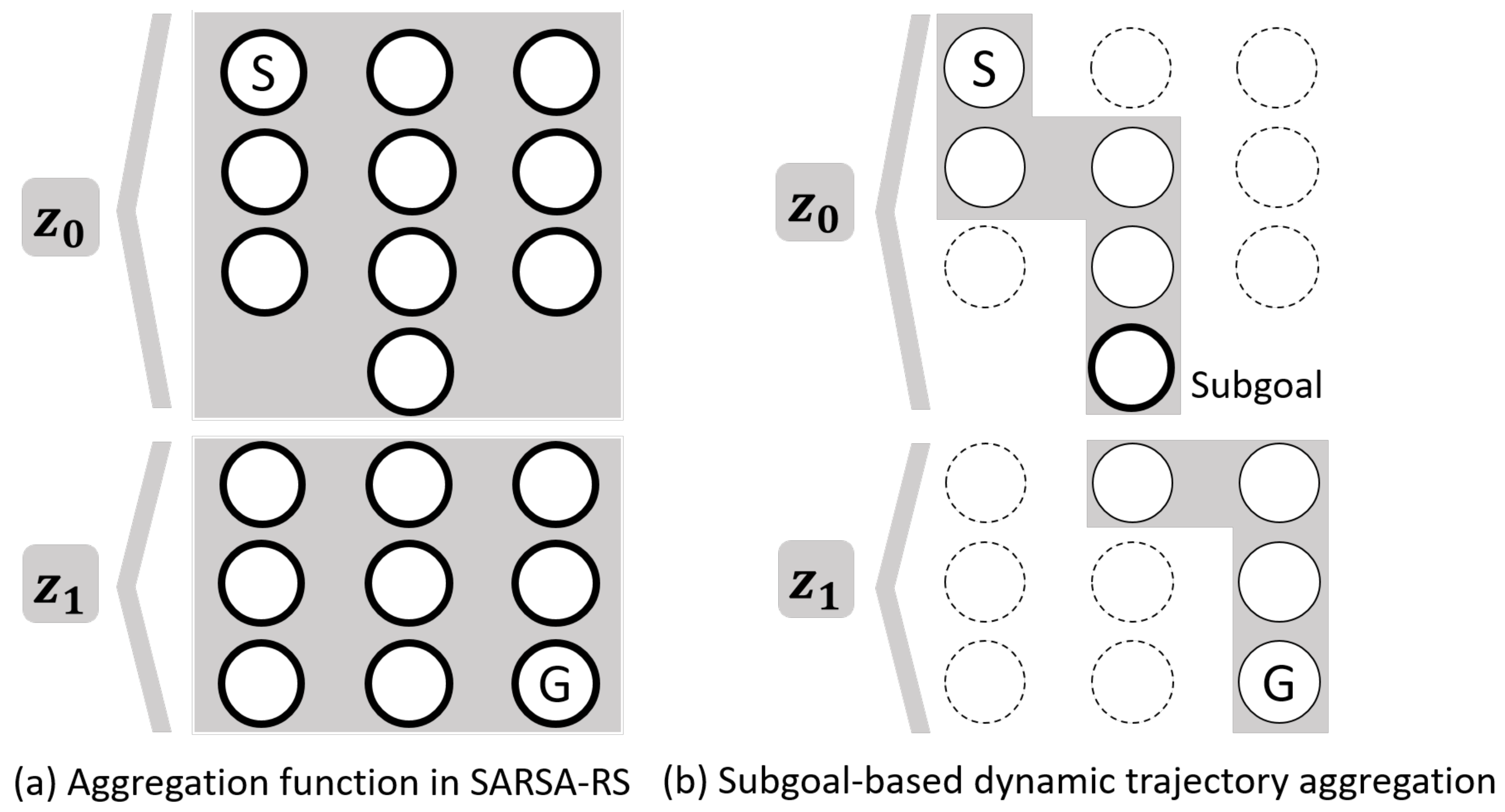}
	\caption{Concept of subgoal-based aggregation.}
	\label{fig:concept}
\end{figure}

We build abstract states to represent the achievement status of a subgoal series. If there are $n$ subgoals, the size of abstract states is $n+1$. The agent is in a first abstract state~$z_0$ before a subgoal is achieved. Then, the abstract state $z_0$ transits to $z_1$ when the subgoal $sg_0$ is achieved. This means the aggregation of episodes until subgoal $sg_0$ transits into $z_0$.  The aggregated episodes change dynamically every trial because of the policy with randomness and learning. As the learning progresses, the aggregated episodes become fixed. The value over abstract states is distributed to the values of states of the trajectory. Note that the trajectories for updating the values are different from those of distributed values. The updated value function is not used for the current trial but for the next trials. An image of dynamic trajectory aggregation is shown in Fig.\ref{fig:concept}.
In the figure, a circle is a state, and the aggregated states are in each gray background area. There are two abstract states in the case of a single subgoal. The bold circles express the states with which the designer deals. The number of bold circles in Fig.~\ref{fig:concept}(b) is much lower than Fig.~\ref{fig:concept}(a). ``S" and ``G" in the circles are a start and a goal, respectively.  Fig.~\ref{fig:concept}(b) shows that the episode is separated into two sub-episodes, and each of them corresponds to the abstract states. 

\subsubsection{Accumulated Reward Function}
We clearly define the reward transformation function~$r_h$ because our method only updates the achievements of subgoals as abstract states. A set of abstract states is part of the semi-Markov decision process~(SMDP)\cite{Sutton:1999:MSF:319103.319108}. The transition between an abstract state and another consists of multiple actions. The value function in SMDP is written as:
\begin{eqnarray}
	V^{g}(z) = E \left\{ \sum_{i=0}^{k-1} \gamma^ir_{t+1+i} + \gamma^k V^{g} (z') | \varepsilon(g, z, t) \right\}
\end{eqnarray}
where $k$ is the duration of the abstract state $z$, and $\varepsilon$ is the event of the aggregation function $g$ being initiated in state $z$ at time $t$. Therefore, we describe this formally as $r_h = \sum_{i=0}^{k-1} \gamma^{i} r_i$, where $k$ is the duration until subgoal achievement. The function accumulates rewards with discount~$\gamma$. Depending on the policy at the time, $k$ is varied dynamically. This follows n-step temporal difference~(TD) learning~\cite{Sutton1998} because there are transitions between an abstract state $z_i$ and another one $z_{i+1}$. 
Algorithm~\ref{alg:subgoal_online_learning} shows the whole process of SARSA-RS with subgoal-based dynamic trajectory aggregation. $\alpha_v$ and $\gamma_v$ are hyper-parameters, that is, the learning rate and discount factor for updating the value function over abstract states.

\begin{algorithm}
	\caption{SARSA-RS with subgoal-based dynamic trajectory aggregation} 
	\label{alg:subgoal_online_learning}
	\KwData{$t=0, V(z; \theta), sg_i \in \{SG, \prec \}$}
	Initialize $\theta$ \\
	$z \leftarrow z_0, i \leftarrow 0, r_h \leftarrow 0$ \\
	Select $a$ by $\pi$ at $s$ \\
	\Repeat{terminal condition}{
		Take $a$ and observe $s'$ and $r$\\
		$z' \leftarrow filter(s')$\\
		$t \leftarrow t+1 $ \\
		$r_h \leftarrow r_h + \gamma^tr$ \\
		\If{$equal(s', sg_{i+1})$}{
			$\delta = r_h + \gamma_v^tV(z'; \theta) - V(z; \theta)$ \\
			$\theta \leftarrow \theta + \alpha_v \delta_\theta \Delta V(z; \theta)$ \\
			$t \leftarrow 0, i \leftarrow i + 1$ \\
		}
		$F(z,z') = \gamma V(z') - V(z)$ \\
		Select $a'$ by $\pi$ \\
		Update value function with $r + F(z,z')$ \\
		$s \leftarrow s'; a \leftarrow a', z \leftarrow z'$
	}
\end{algorithm}

In Algorithm~\ref{alg:subgoal_online_learning}, the method is involved between lines 6-13.  The value function over abstract states is parameterized by $\theta$. If $s'$ equals $sg_{i+1}$, $\theta$ is updated by an approximate multi-step TD method \cite{Sutton1998}, and our method sets the next subgoal $sg_{i+1}$ in lines 9-12.

\section{Experiments} 
In this section, we explain the experiments done to evaluate our method. We used navigation tasks in two domains, four-rooms and pinball, because they are popular problems with discrete/continuous states that have been used in previous studies~\cite{Bacon:2017:OA:3298483.3298491,NIPS2009_3683,Sutton:1999:MSF:319103.319108}. Furthermore, we used a pick and place task with a robot arm with continuous actions\cite{1606.01540}. The navigation task involved finding the shortest path to a goal state from a start state. The pick and place task is to grasp an object and bring it to a target. First, we conducted an experiment to acquire human subgoals. Second, a learning experiment was conducted, in which we compared the proposed method with four other methods for the navigation task. We compared the proposed method with a baseline RL algorithm for the pick and place task. A SARSA algorithm was used for the four-rooms domain, an actor-critic algorithm for the pinball domain, and a DDPG for the pick and place domain. All the experiments were conducted with a PC [Core i7-7700 (3.6GHz), 16GB memory]. We used the same hyper-parameters, $\alpha_v$ and $\gamma_v$, as those of the baseline RL algorithms, respectively.

\subsection{User Study: Human Subgoal Acquisition}
\subsubsection{Navigation Task}
We conducted an online user study to acquire human subgoal knowledge using a web-based GUI. We recruited 10 participants who consisted of half graduate students in the department of computer science and half others(6males and 4females, ages 23 to 60, average of 36.4). We confirmed they did not have expertise on subgoals in the two domains. Participants were given the same instructions as follows for the two domains, and they were then asked to designate their two subgoals both for the four-rooms and pinball domains in this fixed order. The number of subgoals was the same as the hallways in the optimal trajectory for the four-rooms domain.
The instructions explained to the participants what the subgoals were and how to set them. Also, specific explanations of the two task domains were given to the participants. In this experiment, we acquired just {\em two} subgoals for learning since they are intuitively considered easy to give on the basis of the structure of the problems. We considered the two subgoals to be totally ordered ones.

\subsubsection{Pick and Place Task}
A user study for the pick and place task was also done online. Since it was difficult  to acquire human subgoal knowledge with GUI, we used a descriptive answer-type form. We assumed that humans use subgoals when they teach behavior in a verbal fashion. They state not how to move but what to achieve in the middle of behavior. The results of this paper minorly support this assumption. We recruited five participants who were amateurs in the field of computer science(3 males and 2 females, ages 23 to 61, average of 38.4). The participants read the instructions and then typed the answer in a web form. The instructions consisted of a description of the pick and place task, a movie of manipulator failures, a glossary, and a question on how a human can teach successful behavior. The question included the sentence ``Please teach behavior like you would teach your child." This is because some participants answered that they did not know how to teach the robot in a preliminary experiment. We imposed no limit on the number of subgoals.

\subsection{Navigation in Four-Rooms Domain}
The four-rooms domain has four rooms, and the rooms are connected by four hallways. The domain is common for reinforcement learning tasks. In this experiment, learning consisted of a thousand episodes. An episode was a trial run until an agent reached a goal state successfully or when a thousand state-actions ended in failure. A state was expressed as a scalar value labeled through all states. An agent could select one of four actions: up, down, left, and right. The transition of a state was deterministic. A reward of +1 was generated when an agent reached a goal state. The start state and goal state were placed at fixed positions. An agent repeated the learning 100 times. The learning took several tens of seconds.

\subsubsection{Experimental Setup}
We compared the proposed reward shaping with human subgoals~(HRS) with three other methods. They were a SARSA algorithm~(SARSA)~\cite{Sutton1998}, the proposed reward shaping with random subgoals~(RRS), and naive subgoal reward shaping~(NRS). 
SARSA is a basic reinforcement learning algorithm. We used SARSA as a baseline algorithm and implemented the other two methods with it. RRS used two randomly selected states as subgoals from the whole state space. NRS is based on potential-based reward shaping The potential function $\Phi(s)$ outputs a scalar value $\eta$ just when an agent has visited a subgoal state. The potential function is written formally as follows.
\begin{eqnarray}
	\Phi(s) = \left \{\begin{array}{ll}
		\eta&s = sg \\
		0&s\neq sg
	\end{array}
	\right.
\end{eqnarray}
Informally, NRS shapes rewards of $\eta$ only generated for subgoals with potential-based reward shaping. The two differences from our method are that NRS has a fixed potential, and the positive potential only for the subgoals. The reward shaping methods were given ordered subgoals or aggregation of states in advance. We set the learning rate for SARSA to 0.01, the discount rate for SARSA to 0.99, and $\eta$ to 1.0. The policy was a softmax. We chose $\eta$ to be 1 so that it would be the same value as the goal reward after grid search on grids of 1, 10, and 100.\par
We evaluated the learning performance with the time to threshold and the asymptotic performance~\cite{Taylor:2009:TLR:1577069.1755839} in terms of the learning efficiency of transfer learning for reinforcement learning. We explain the definitions of the measurements in this experiment. The time to threshold was the number of episodes required to get below a pre-defined threshold of steps. The asymptotic performance was the final performance of learning. We used the average number of steps between 990 and 1000 episodes.
\subsubsection{Experimental Results}
\begin{figure}
	\centering
	\includegraphics[width=\linewidth]{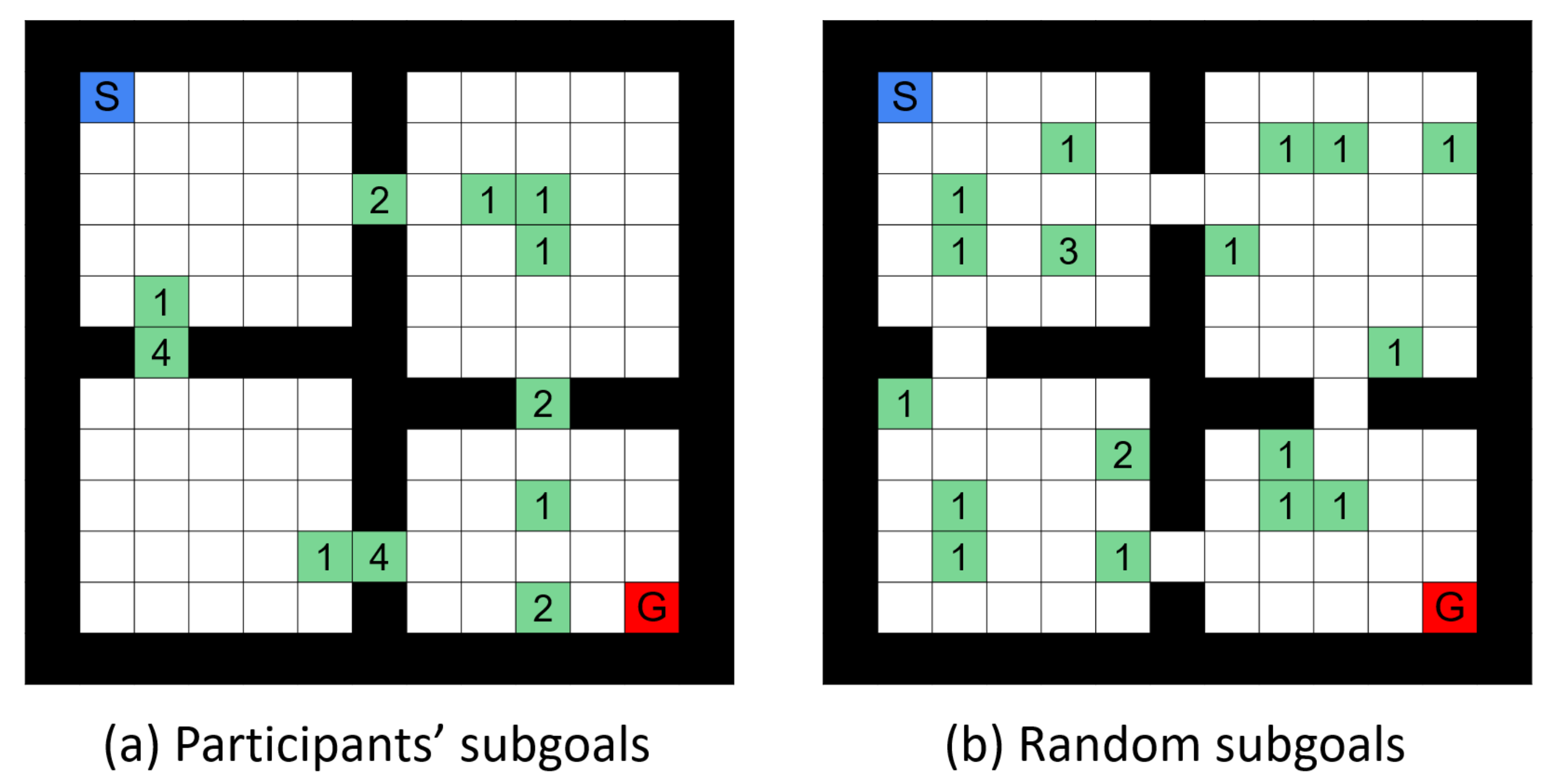}
	\caption{Subgoal distribution for four-rooms domain.}
	\label{fig:subgoal-dist}
\end{figure}

Fig.~\ref{fig:subgoal-dist} shows the subgoal distribution acquired from the ten participants and from the random subgoals generated for the four-rooms domain. 
\begin{figure}
	\centering
	\includegraphics[width=\linewidth]{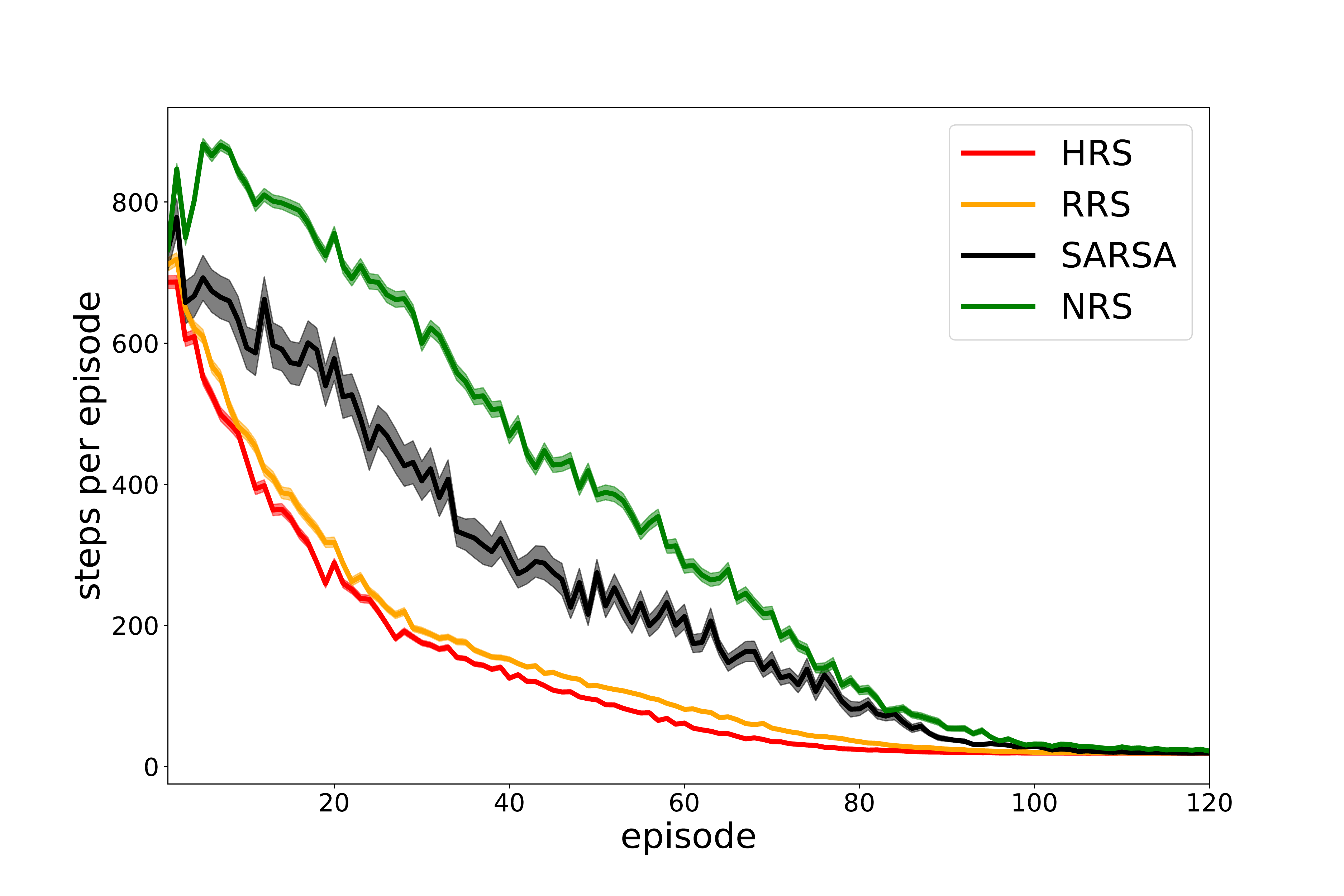}
	\caption{Learning curves compared among four methods.}
	\label{fig:learning-curves}
\end{figure}
In Fig.~\ref{fig:subgoal-dist}, the color of the start, goal, and subgoal cells are red, blue, and green, respectively. 
The number in a cell is the frequency at which the participants selected the cell as a subgoal. These subgoals included totally and partially ordered subgoals. As shown in Fig.~\ref{fig:subgoal-dist}(a), participants tended to set more subgoals in the hallway compared with random subgoals  [Fig.~\ref{fig:subgoal-dist}(b)]. 
Next, we show the results of the learning experiment. Fig.~\ref{fig:learning-curves} shows the learning curves of our proposed method and the four other methods. The standard errors also are shown in this figure.
We plotted HRS with an average totaling 1000 learnings over all participants. RRS were also averaged by 1000 learnings over 10 patterns. NRS had almost the same conditions as HRS. SARSA was averaged by 10,000 learnings. HRS had the fewest steps for almost all episodes. The results of NRS demonstrated the difficulty with transformation from subgoals into an additional reward function. We also performed an ANOVA among the four methods. We set the thresholds to 500, 300, 100, and 50 steps in terms of the time to threshold. Table~\ref{tab:mean-sd} shows the mean episodes, the standard deviations and the results of the ANOVA and the sub-effect tests for the compared methods for each threshold step.

\begin{table}[tb]
	\caption{Mean and standard deviation. Mean~(S.D.) and results of ANOVA and sub-effect tests in episodes to threshold steps in four-rooms domain.}
	\label{tab:mean-sd}
	\begin{center}
		\begin{tabular}{rcccc}
			\toprule
			Thres.&HRS&RRS&SARSA&NRS\\
			\midrule
			500 & 2.68(1.86) &  2.91(2.10) & 3.93(3.07) & 5.78(6.39)\\
			300 & 5.06(3.05) & 5.60(3.61) & 6.68(4.29) & 10.4(9.17)\\
			100 & 17.3(7.92) & 18.6(7.89) & 26.4(11.0) & 38.3(47.1)\\
			50 & 33.0(10.2) & 36.3(11.0) & 51.8(16.2) & 59.6(45.8)\\
			\bottomrule
		\end{tabular}
	\end{center}
\end{table}

\begin{table}[tb]
	\caption{Summary of ANOVA and sub-effect tests in episodes to threshold steps. }
	\label{tab:time-to-threshold}
	\begin{center}
		\begin{tabular}{rccc}
			\toprule
			\hspace{-1em}Thres.&\hspace{-1em}HRS\verb|<|RRS&\hspace{-1em}\{HRS,RRS\}\verb|<|SARSA&\hspace{-1em} \{HRS,RRS,SARSA\}\verb|<|NRS\\
			\midrule
			500 & n.s. & * & *\\
			300 & * & * & *\\
			100 & n.s. & * & *\\
			50 & * & * & *\\
			
			\bottomrule
		\end{tabular}
	\end{center}
\end{table}

As shown in Table~\ref{tab:mean-sd}, HRS shortened the required time to approximately 20 episodes for reaching the 50 steps, which was better than the performance of RRS. We did not find a statistically significant difference between HRS, RRS, SARSA, and NRS in terms of asymptotic performance. Our method made the learning faster than the baseline method, and human subgoals lead to better performance than random ones.

\subsection{Navigation in Pinball Domain}
The navigation task in the pinball domain involves moving a ball to a target by applying force to it. The pinball domain is difficult for humans because delicate control of the ball is necessary. This delicate control is often required in the control domain. Since humans only provide states, ordered subgoals are more tractable than nearly optimal trajectories in such domains. \par
The difference with the four-rooms domain is the continuous state space over the position and velocity of the ball on the x-y plane. An action space has five discrete actions, four types of force and no force.  In this domain, a drag coefficient of 0.995 effectively stops the ball from moving after a finite number of steps when the no-force action is chosen repeatedly; collisions with obstacles are elastic. The four types of force were up, down, right, and left on a plane. Actions were randomly chosen at 10\%. An episode terminated with a reward of +10000 when the agent reached the goal. Interruption of any episode occurred when an episode took more than 10,000 steps. The radius of a subgoal was the same as a goal.

\subsubsection{Experimental Setup}
We compared HRS with AC, RRS, and NRS in terms of learning efficiency with the time to threshold and the asymptotic performance. We defined this threshold in the domain as the number of episodes required to reach the designated number of steps. The asymptotic performance was the average number of steps between 190 and 200 episodes. A learning consisted of 200 episodes at most. All methods learned a total 100 times from scratch. HRS, RRS, and NRS performed ten learnings with ten patterns. HRS and NRS used two ordered subgoals provided by ten participants. RRS used two ordered subgoals generated randomly. We used the results to evaluate the learning efficiency.
The learning took several tens of minutes. A subgoal had only a center position and a radius. The radius was the same as that of the target. A subgoal was achieved when the ball entered the circle of it at any velocity. We used AC as the critic with linear function approximation over a Fourier basis~\cite{Konidaris11a} of order three. We used the linear function as the actor, and a softmax policy decided an action. The learning rates were set to 0.01 for both the actor and the critic. The discount factor was set to 0.99. The $\eta$ of NRS was 10,000 so as to be the same value as the goal reward.

\subsubsection{Experimental Results}
\begin{figure}[tb]
	\centering
	\includegraphics[width=\linewidth]{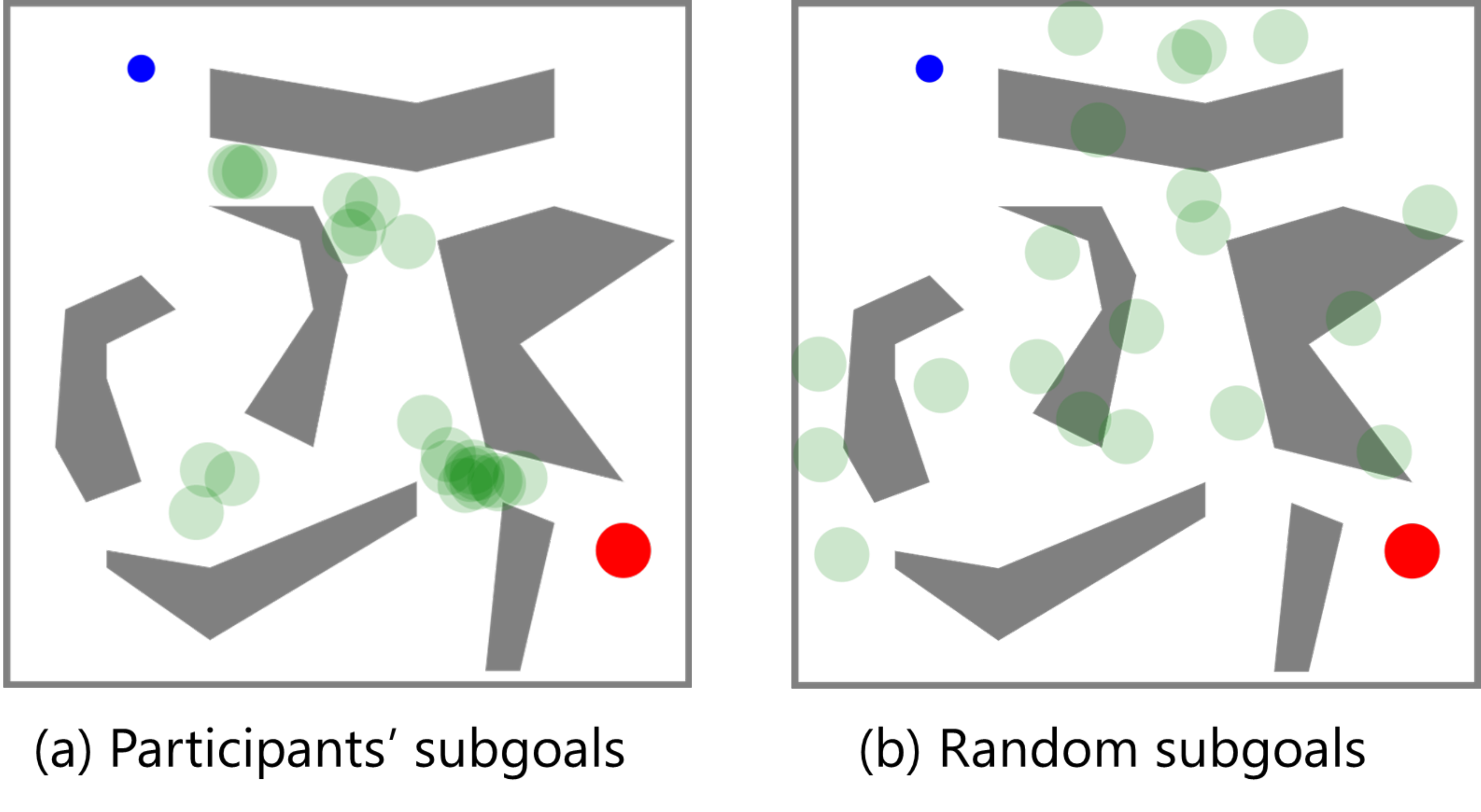}
	\caption{Subgoal distribution in pinball domain.}
	\label{fig:subgoal-pinball}
\end{figure}

Fig.~\ref{fig:subgoal-pinball} shows the subgoal distribution acquired from participants and from the random subgoals generated for the pinball domain. In this figure, the color of the start point, the goal, and subgoals are red, blue, and green, respectively. As shown in Fig.~\ref{fig:subgoal-pinball}, participants focused on four regions of branch points to set subgoals in comparison with random subgoals. 

\begin{figure}[tb]
	\centering
	\includegraphics[width=\linewidth]{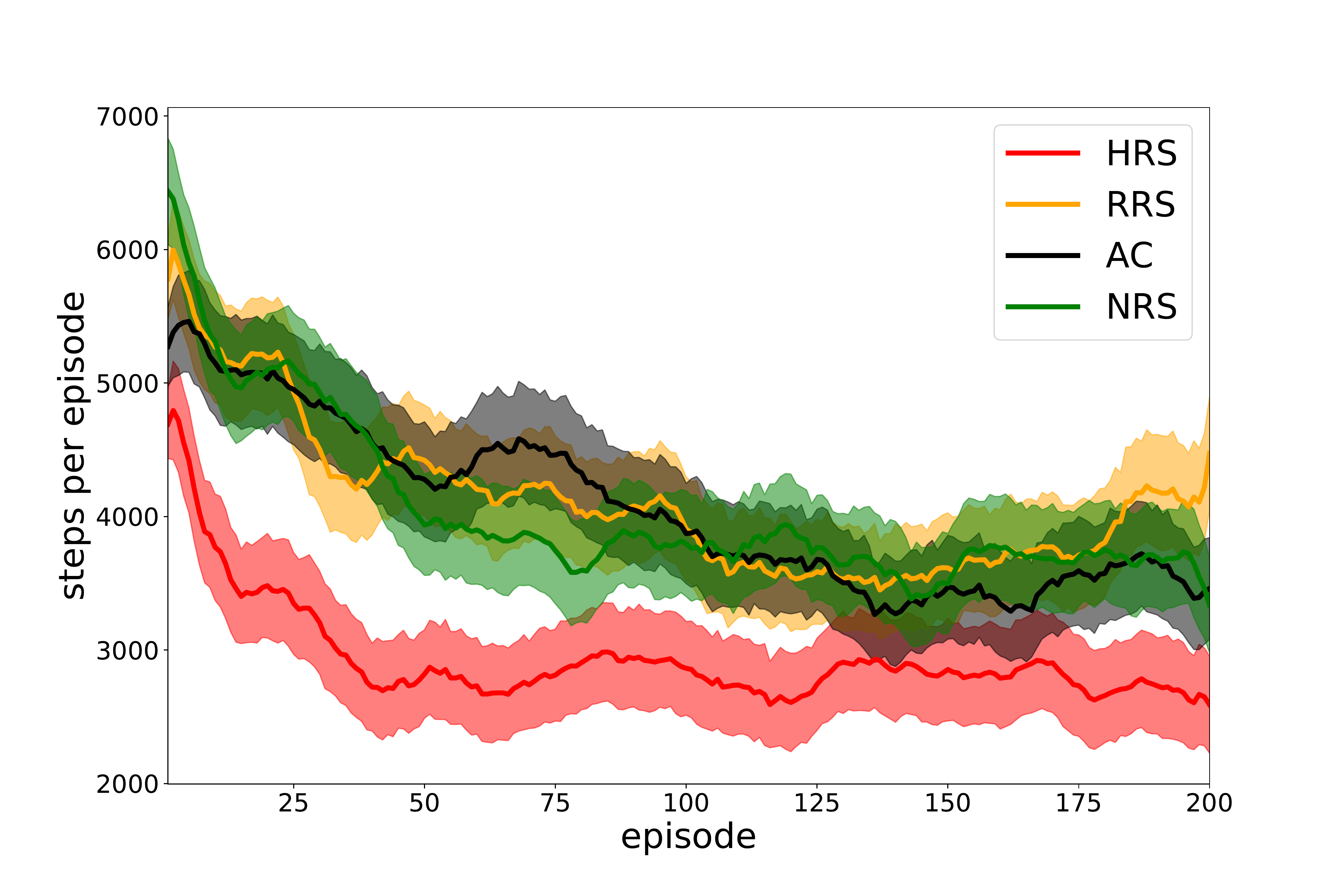}
	\caption{Learning curves in pinball domain.}
	\label{fig:lc-pinball}
\end{figure}

Fig.~\ref{fig:lc-pinball} shows the learning curves of HRS, RRS, AC, and NRS. The learning indicator was the average number of steps per episode over learning 100 times. It took an average shift of ten episodes. The standard errors also are shown in this figure.
As shown, HRS performed better than all other methods. RRS and NRS seemed to be almost the same as AC. We evaluated the learning efficiency by using the time to threshold and the asymptotic performance. We used each learning result smoothed using a simple moving average method with the number of time periods at ten episodes. 
We performed an ANOVA to determine the difference among the four methods, HRS, RRS, AC, and NRS. The Holm-Bonferroni method was used for a sub-effect test. We set the thresholds to 3000, 2000, 1000, and 500 steps in terms of the time to threshold. Table~\ref{tab:summary-pinball} shows the mean episodes, the standard deviations and the results of the ANOVA and the sub-effect tests for the compared methods for each threshold step.
\begin{table}[tb]
	\caption{Mean and standard deviation:Mean~(S.D.) in pinball domain.}
	\label{tab:summary-pinball}
	\begin{center}
		\begin{tabular}{rcccc}
			\toprule
			Thres.~&~HRS~&~RRS~&~AC~&~NRS~\\
			\midrule
			~3000~~& 21.2(39.8) & 34.6(48.4) & 51.7(67.1) & 40.0(50.7)\\
			~2000~~& 26.8(40.4) & 48.7(54.2) & 70.2(77.8) & 59.0(59.0)\\
			~1000~~& 56.7(58.5) & 89.1(67.9) & 101(71.4) & 93.0(70.10)\\
			~500~~& 115.3(66.4) & 147.3(64.7) & 163(54.8) & 157(61.6)\\
			\bottomrule
		\end{tabular}
	\end{center}
\end{table}


\begin{table}[tb]
	\caption{Summary of ANOVA and sub-effect tests in episodes to threshold steps in pinball domain.}
	\label{tab:time-to-threshold-pinball}
	\begin{center}
		\begin{tabular}{rcc}
			\toprule
			Thres. & HRS\verb| < |RRS&HRS\verb| < |$\{$RRS, AC$\}$\\
			\midrule
			3000 & n.s. & * \\
			2000 & * & * \\
			1000 & * & * \\
			500 & * & * \\
			\bottomrule
		\end{tabular}
	\end{center}
\end{table}

From Table~\ref{tab:summary-pinball}, the difference between HRS and the other three methods in terms of reaching 500, 1000, and 2000 steps per episode was statistically significant. There were statistically significant differences between HRS and both RRS and AC in terms of reaching 3000 steps. There were no other significant differences. This means that HRS learned faster until reaching steps than RRS, AC, and NRS. 
There was only statistically significant difference between HRS and RRS in terms of asymptotic performance. From these results, we found human ordered subgoals to be more helpful for our proposed method than random ordered subgoals in the pinball domain. We acquired similar results for the four-rooms domain. 

\subsection{Pick and Place Task}
We used a fetch environment based on the 7-DoF Fetch robotics arm of OpenAI Gym\cite{1606.01540}. In the pick and place task, the robotic arm learns to grasp a box and move it to a target position\cite{DBLP:journals/corr/abs-1802-09464}. We converted the original task into a single-goal reinforcement learning framework because potential-based reward shaping does not cover the multi-goal framework\cite{Ng+HR:1999}. The dimension of observation is larger than the previous navigation task, and the action is continuous. An observation is 25-dimensional, and it includes the Cartesian position of the gripper and the object as well as the object's position relative to the gripper. The reward function generates a reward of -1 every step and a reward of 0 when the task is successful. In \cite{DBLP:journals/corr/abs-1802-09464}, the task is written about in detail.

\subsubsection{Experimental Setup}
We compared HRS with NRS, RRS, and DDPG~\cite{journals/corr/LillicrapHPHETS15} in terms of learning efficiency with the time to threshold and asymptotic performance. HRS, NRS, and RRS used DDPG as the base. We defined this threshold in the task as the number of epochs required to reach the designated success rate. The asymptotic performance was the average success rate between 190 and 200 epochs. Ten workers stored episodes and calculated the gradient simultaneously in an epoch. HRS and NRS used ordered subgoals provided by five participants. RRS used subgoals randomly generated. The learning in 200 epochs took several hours. We used an OpenAI Baselines\cite{baselines} implementation for DDPG with default hyper-parameter settings. We built the hidden and output layers of the value network over abstract states with the same structure as the q value network. We excluded the action from the input layer. The input of the network is only the observation on subgoal achievement, and the network learns from the discount accumulated reward until subgoal achievement. A subgoal is defined from the information in the observation. We set the margin to $\pm{0.01}$ to loosen the severe condition to achieve subgoals.

\subsubsection{Experimental Results}
\begin{figure}[tb]
	\centering
	\includegraphics[width=\linewidth]{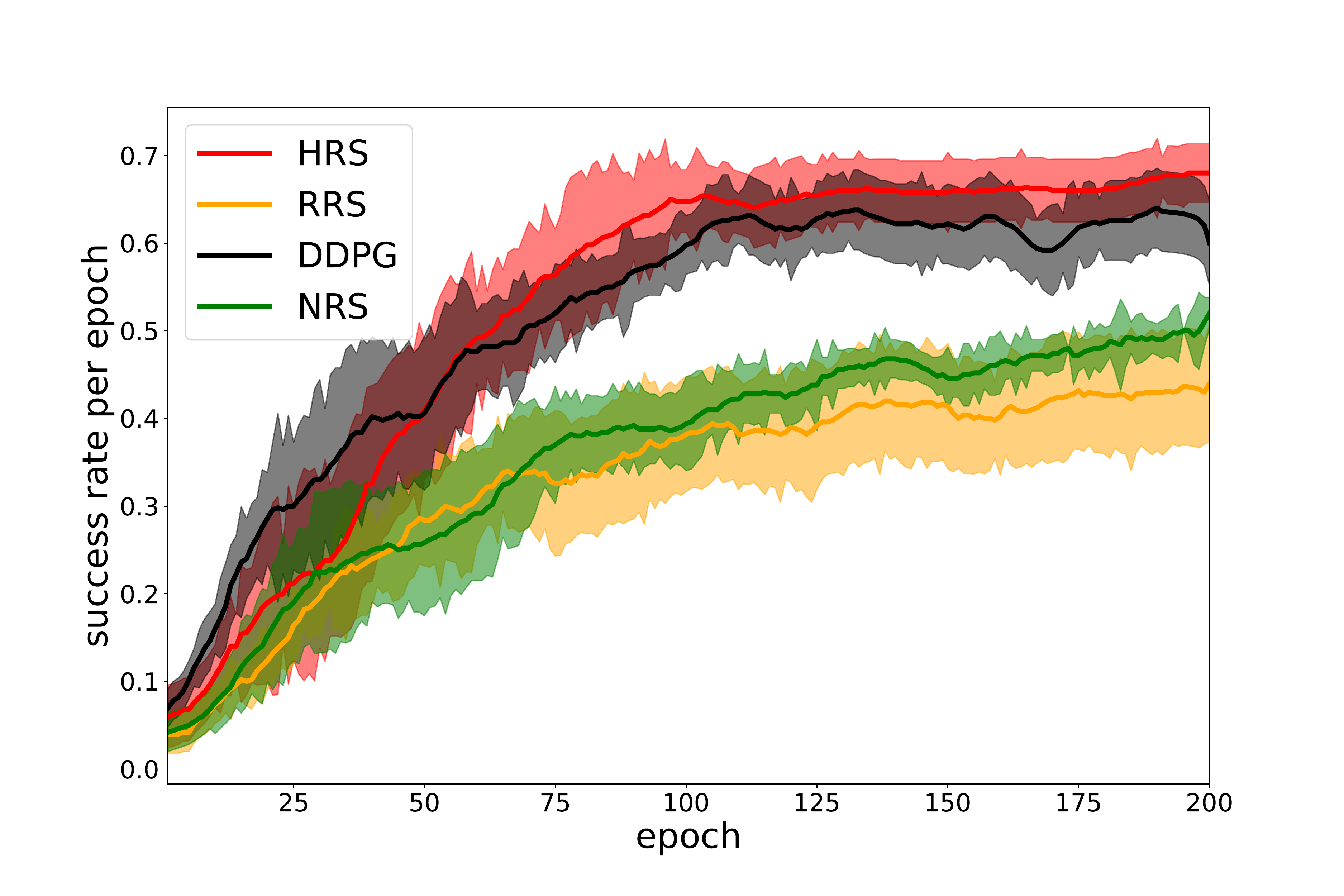}
	\caption{Learning curves in pick and place task.}
	\label{fig:lc-robotics}
\end{figure}

All five participants determined the subgoal series, the first subgoal was the location available to grasp the object, and the second subgoal was grasping the object. We used the subgoal series for the input of our method Fig.~\ref{fig:lc-robotics} shows the learning curves of HRS, RRS, NRS, and DDPG. 

As shown as Fig.~\ref{fig:lc-robotics}, the results were averaged across five learnings, and the shaded areas represent one standard error. The random seeds and the locations of the goal and object were varied every learning. HRS worked more effectively than DDPG, especially after about the 75th epoch. NRS had the worst performance through almost all epochs.
The mean difference was 30 epochs at a time to threshold of 0.6. This means that our method decreased the number of epochs by 30 to reach the success rate of 0.6 from the DDPG. The NRS and RRS could not achieve the success rate of 0.6. The asymptotic performance of HRS, DDPG, NRS, and RRS are 0.67, 0.63, 0.49, and 0.43 respectively. We confirmed that HRS had the highest asymptotic performance and the fastest achievement at the 0.6 success rate.

\section{Discussion}
There was a small difference between HRS and RRS in terms of navigation in the four-rooms domain as shown in Fig.~\ref{fig:learning-curves}. In comparison, the difference was similar to AC and NRS for the pinball domain as shown in Fig.~\ref{fig:lc-pinball}. Approximately 65\% of states generated randomly were in the optimal trajectory for the four-rooms domain. The pinball task had approximately 20\% of states generated randomly in it. The random subgoals were better for the four-rooms domain than in the pinball domain. This is because the four-rooms domain might have more states in an optimal trajectory than the pinball domain. We think that the smaller difference was caused by the characteristics of the four-rooms task, for which most states were in the optimal trajectory. \par
Potential-based reward shaping keeps a policy invariant from the transformation of a reward function. From the experimental results of the pinball domain, the asymptotic performances of HRS were statistically significantly different from RRS. There was no significant difference in the four-rooms domain. As shown in Fig.~\ref{fig:learning-curves}, the performance was clearly asymptotic at the 121st of 1000 episodes. In contrast, it was not asymptotic at the 200th episode in Fig.~\ref{fig:lc-pinball}. Since our method is based on potential-based reward shaping, the RRS converges to the same performance as HRS if learning continues. \par
We compared our method with SARSA-RS using state aggregation. The comparison is not fair because the amount of domain knowledge is different. Our method needs only several states as subgoals, whereas the state aggregation needs the mapping from states into abstract states. However, the comparison is useful to understand the performance of our method in detail. For the four-rooms domain having a discrete state space, state space aggregation is easily given such that an abstract state is a room. In the pinball domain and pick and place task having continuous state space, it is difficult to obtain a human's mapping function. This is why we got the abstract state space by discretization. The number of both abstract states was three to align SARSA-RS with HRS. The performance of our method was lower than the state aggregation for the four-rooms and pinball domains. In comparison, our method outperformed the state aggregation in the pick and place task. The results may show that state aggregation does not work in a high-dimensional environment, but our method works well. \par
The value function over abstract states was initialized by zero in the experiments. As shown in the experimental results, our method improved the learning efficiency in the middle of learning, but could not speed up RL in the beginning. This is because the shaping was zero by the initial potentials, and did not work well in the begging. Since non-zero initialization can shapes the reward in the begging, it might speed up RL. However, the best way to initialize the value function is not clear. This is an open question.\par
The limitation of providing subgoals is that there is no established methodology, and it may depend on the choices each individual intuitively makes. Future research is hence needed to define subgoals clearly. For the four-rooms domain, as shown in Fig.~\ref{fig:subgoal-dist}, almost all of the subgoals were scattered within the right-top and right-bottom rooms. From this, we think that many participants tended to consider the right-bottom path as the shortest one.  Additionally, there was an interesting observation in that a half of the participants set a subgoal in the {\em hallways}. This may mean humans abstractly have a common methodology and preference for giving subgoals. It is necessary to systematically conduct a user study to make this clear. \par
Providing subgoals is more useful than optimal trajectories when the task requires robot-handling skills. We are interested in cognitive loads when a human teaches behaviors to a learning algorithm. There are three points are left as open problems: choosing suitable tasks to provide subgoals, measuring the quantitative difference in cognitive load among the types of provided human knowledge, and developing a graphical user interface~(GUI) for teaching by subgoals. We consider tasks with perceptual structures such as navigation in four-rooms to be suitable for providing subgoals. The four-rooms domain is a grid, and the structure is explicitly clear, so hallways between rooms tends to be selected. If the task has a single room, participants would be confused and unsure of where to select subgoals. Tasks without perceptual structures may be suitable for providing optimal trajectories. The GUI is significant to both teachers and agents. The cognitive load of teachers may decrease, and the appropriate subgoals can be acquired to accelerate learning. The agent needs to have interpretability in regards to its behaviors so that human can acquire the desired information for efficiency. We will consider incorporating the XAI approach \cite{molnar2019} into the GUI. 
\section{Conclusion}
In reinforcement learning, learning a policy is time-consuming. We aim for accelerating learning with reward transformation based on human subgoal knowledge. Although SARSA-RS incorporating state aggregation information into rewards is helpful, humans rarely deal with all states in an environment with high-dimensional observations. We proposed a method by which a human deals with several characteristic states as a subgoal. We defined a subgoal as the goal state in one of the sub-tasks into which a human decomposes a task. The main part of our method is the dynamic trajectory aggregation with subgoal series into abstract states. The method works well with an accumulated reward function in the environment. The accumulated reward function returns rewards of n-step transitions. We collected ordered subgoals from participants and used them for evaluation. We evaluated navigation for four-rooms, pinball, and a pick and place task. The experimental results revealed that our method with human subgoals enabled faster learning compared with the baseline method, and human subgoal series were more helpful than random ones. We could apply the SARSA-RS with our method to an environment with high-dimensional observations, and learning was clearly accelerated. Future work involves analyzing the characteristics of human subgoals to clearly define the subgoals humans provide.

\bibliographystyle{IEEEtranS.bst}
\bibliography{Bibliography.bib}

\end{document}